\documentclass{article}
\usepackage{ijcai22}

\usepackage{times}
\usepackage{soul}
\usepackage{url}
\usepackage[hidelinks]{hyperref}
\usepackage[utf8]{inputenc}
\usepackage[small]{caption}
\usepackage{graphicx}
\usepackage{amssymb}
\usepackage{amsfonts}
\usepackage{amstext}
\usepackage{amsmath}
\usepackage{amsthm}
\usepackage{booktabs}
\usepackage{algorithm}
\usepackage{algorithmic}
\usepackage{makecell}
\usepackage{multirow}
\usepackage{transparent}
\usepackage{tikz}

\usepackage{color}
\newcommand{\todo}[1]{\textbf{\textcolor{red}{#1}}}

\usepackage{tabularx}
\usepackage{etoolbox,siunitx}
\robustify\bfseries
\usepackage[font=scriptsize,skip=2pt]{subcaption}

\newcommand{\rf}{\textit{ref}.}
\newcommand{\ie}{\textit{i}.\textit{e}.}
\newcommand{\eg}{\textit{e}.\textit{g}.}
\newcommand{\vs}{\emph{vs}.}

\usepackage[capitalize]{cleveref}
\crefname{section}{Sec.}{Section}

\usepackage{footnote}
\usepackage{footmisc}
\renewcommand{\thefootnote}{\alph{footnote}}
\newcommand{\astfootnote}[1]{%
\let\oldthefootnote=\thefootnote%
\setcounter{footnote}{0}%
\renewcommand{\thefootnote}{\fnsymbol{footnote}}%
\footnote{#1}%
\let\thefootnote=\oldthefootnote%
}

\title{RAFT-MSF: Self-Supervised Monocular Scene Flow using Recurrent Optimizer}

\author{
Bayram Bayramli$^1$
\and
Junhwa Hur$^2$\astfootnote{Work done prior to joining 42dot.ai}
\and
Hongtao Lu$^{1}$
\affiliations
$^1$Shanghai Jiao Tong University\\
$^2$42dot.ai\\
}
\begin{document}

\maketitle

\begin{abstract}
    \if False
Learning 3D motion field from monocular cameras in a un/self-supervised fashion still remains a challenging task. Due to the absence of having widespread ground truth labels and the sensitivity of the objective functions, the performance of the current self-supervised methods is lagging behind those of semi-supervised or fully supervised methods. In this paper, we introduce self-supervised monocular scene flow network to iteratively update 3D motion fields and disparity maps simultaneously. Specifically, our approach integrates RAFT, a supervised optical flow model, with self-supervised proxy losses, and with a new architecture that suits not only 3D motion field but also disparity estimation together. Furthermore, we also propose gradient detaching strategy, and disparity initialization technique to improve on state-of-the-art methods. Extensive experiments validate that our method achieves state-of-the-art performance among self-supervised monocular scene flow methods. We achieve a scene flow error of $30.97\%$, outperforming the best-published method, $47.05\%$, on two-view input. Code will be available.
\todo{not clear about the metric/unit of the numbers. better to say: improve the accuracy by xx\% ...}
\fi

Learning scene flow from a monocular camera still remains a challenging task due to its ill-posedness as well as lack of annotated data.
Self-supervised methods demonstrate learning scene flow estimation from unlabeled data, yet their accuracy lags behind (semi-)supervised methods.
In this paper, we introduce a self-supervised monocular scene flow method that substantially improves the accuracy over the previous approaches.
Based on RAFT, a state-of-the-art optical flow model, we design a new decoder to iteratively update 3D motion fields and disparity maps simultaneously.
Furthermore, we propose an enhanced upsampling layer and a disparity initialization technique, which overall further improves accuracy up to 7.2\%.
Our method achieves the state-of-the-art accuracy among all self-supervised monocular scene flow methods, improving accuracy by 34.2\%.
Our fine-tuned model outperforms the best previous semi-supervised method with 228 times faster runtime.
Code will be publicly available.
\end{abstract}

\section{Introduction}

%Autonomous navigation systems such as self-driving cars need to be able to move safely in dynamic environments. The 3D motion of the dynamic scene is required to facilitate autonomous agents to navigate securely. Scene flow estimation is the task of learning 3D motion field of points in the world space, equivalently optical flow is the 2D pixelwise displacement field of points in an image plane \cite{Vedula2005ThreedimensionalSF}. Plenty of approaches have been proposed to estimate scene flow from RGB-D videos, stereo videos  \cite{Teed2021RAFT3DSF}, and point clouds \cite{Gu2019HPLFlowNetHP}. However, each setting has its own unique limitations including costly sensing devices, calibration for stereo rig, and accommodating only indoor environments. In this paper, our main focus is learning scene flow from monocular setting due to an easy access to monocular videos on Internet and low-cost configuration of monocular cameras.

For safe navigation, autonomous systems (\eg, self-driving cars, robots) need to comprehend the 3D motion of dynamic scenes, including multiple moving objects.
Scene flow estimation is the task of estimating the 3D motion of 3D points in the world coordinate \cite{Vedula2005ThreedimensionalSF}. Recent state-of-the-art methods \cite{Wang2020FlowNet3DGL,Teed2021RAFT3DSF} estimates scene flow from 3D points or RGB-D cameras in a supervised manner. However, supervised learning requires annotated training datasets. It is expensive to obtain such datasets because labeling the motion of every point in the real world is highly laborious. Therefore, supervised learning-based method primarily used large scale synthetic datasets for training and real-world dataset for fine-tuning only. Hence, such expensive requirements restrict the effectiveness of supervised methods in real-world setting. Moreover, using different types of input sensors (\eg, RGB-D videos, stereo videos  \cite{Teed2021RAFT3DSF}, and point clouds \cite{Wang2020FlowNet3DGL}), brings further limitations for each sensor configuration(\eg, indoor usage only, calibration for stereo rig, costly LiDAR).

\begin{figure}[t]
\centering
\subcaptionbox{Overlayed input images}{\includegraphics[width=0.49\linewidth]{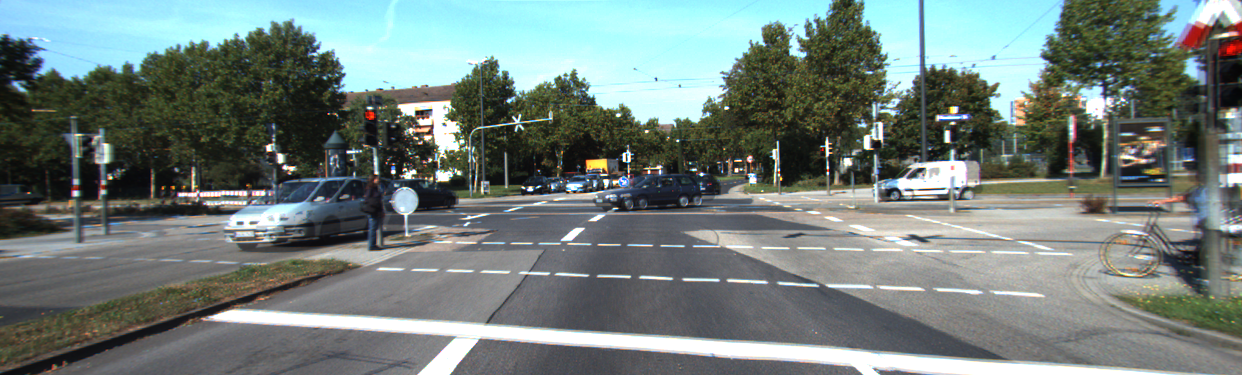}}
\subcaptionbox{Depth map}{\includegraphics[width=0.49\linewidth]{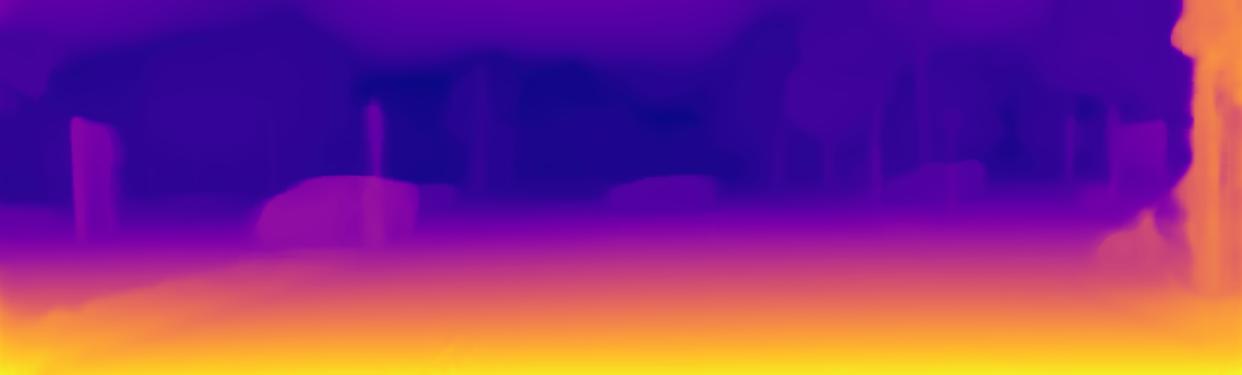}} \\
\subcaptionbox{Optical flow}{\includegraphics[width=0.49\linewidth]{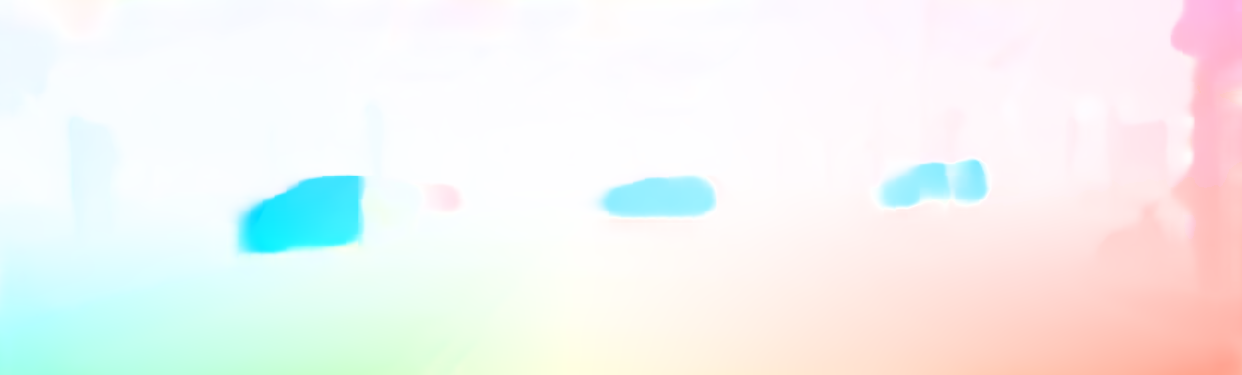}}
\subcaptionbox{Scene flow}{\includegraphics[width=0.49\linewidth]{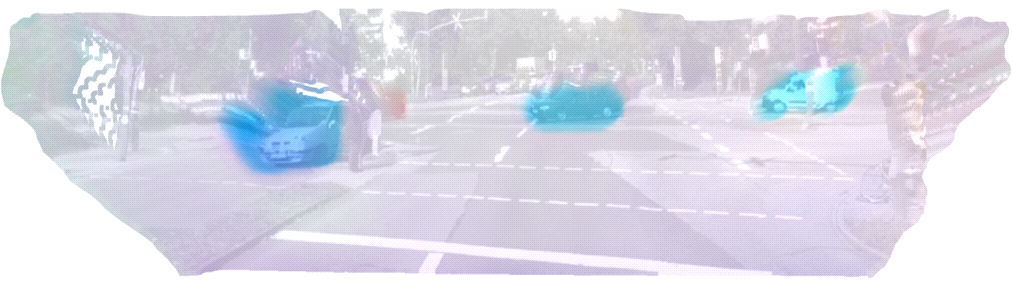}} \\
\caption{Results of our monocular scene flow on KITTI dataset. Given \textit{(a)} two consecutive monocular images, our method outputs \textit{(b)} depth, \textit{(c)} optical flow, and \textit{(d)} scene flow together.}
\label{fig:front_page}
\end{figure}

To overcome the above mentioned limitations, we focus on scene flow estimation using a monocular camera in a self-supervised fashion, which is a simple, low-cost setup that resolves those limitations.
Scene flow estimation from monocular images is a challenging, ill-posed task due to depth ambiguity. Several self-supervised multi-task learning approaches \cite{Yang2018EveryPC,Luo2020EveryPC,Zou2018DFNetUJ} demonstrate the possibility of learning scene flow from only monocular video frames, yet their complicated training pipelines, as well as architectures, limit the practicality of their methods.
Recently, a couple of methods \cite{Hur2020SelfSupervisedMS,Hur2021SelfSupervisedMM} propose more effective approaches to estimate monocular scene flow: directly estimating scene flow from a single network, requiring simple training schedules, and substantially improving the accuracy.
Nevertheless, their accuracy (even after finetuning with available ground truth) still falls behind semi-supervised methods.

%In this work, our main motivation is to show that a simple, trick-free network for self-supervised monocular scene flow can achieve superior results in both scene flow and depth estimation compared to multifarious self-supervised and semi-supervised methods. Therefore, we introduce a new robust architecture for monocular scene flow, named RAFT-MSF. The overall architecture of our network is inspired from RAFT \cite{Teed2020RAFTRA}, a supervised optical flow model that extracts features from two consecutive images, and builds a 4D correlation volume, then iteratively updates 2D motion field based on GRU units. First, we adapt the RAFT model to self-supervised learning with proxy loss functions. We build a new update block that can iteratively estimate scene flow and disparity simultaneously, different from RAFT where it only updates a 2D displacement. We demonstrate that a simple adaptation of the RAFT model fails to learn scene flow and disparity together, however with our new update block we can outperform the previous state-of-the-art methods \cite{Hur2020SelfSupervisedMS,Hur2021SelfSupervisedMM}. Additionally, we introduce a disparity initialization technique on top of the feature extraction model instead of initializing it to zeros. This further improves the ability to learn better disparity and contributes to overall scene flow performance. We also propose a detaching strategy on learnable upsampling module to obviate checkerboard artifacts and stabilize the training.

In this paper, we introduce a new advanced baseline for self-supervised monocular scene flow that achieves superior results in both scene flow and depth over previous monocular scene flow methods, an example of our results is shown in \cref{fig:front_page}. Inspired by RAFT \cite{Teed2020RAFTRA}, a state-of-the-art optical flow architecture, we propose \textbf{RAFT-MSF} that iteratively and residually updates 3D scene flow and depth using a GRU (Gated Recurrent Unit), given a 4D correlation pyramid.
We additionally propose a gradient detach as well as disparity initialization technique that suits for our monocular scene flow task on the RAFT backbone architecture, which overall improves the results both qualitatively and quantitatively.
Especially the gradient detaching technique on learnable upsampling module obviates checkerboard artifacts that vanilla RAFT backbone architecture presents in a self-supervised learning setup.

Our method achieves the state-of-the-art accuracy for self-supervised monocular scene flow.
On KITTI 2015 Train, our method reduces the scene flow of our direct competitor \cite{Hur2020SelfSupervisedMS}, from 47.05 \% (two-view) to 30.97\%, while even outperforming a multi-view method \cite{Hur2021SelfSupervisedMM} (39.82\%).
Our method also achieves superior results in monocular depth estimation despite using a simple network. 
Finally, our fine-tuned version with 200 annotated pairs also outperforms semi-supervised monocular scene flow methods \cite{Brickwedde2019MonoSFMG,Schuster2020MonoCombAS} to date while demonstrating real-time performance.
\section{Related Work}
\paragraph{Scene Flow.}
%Estimating the three-dimensional motion of a field is a longstanding task in computer vision that has been approached in various ways. \cite{Vedula2005ThreedimensionalSF} first introduced the scene flow estimation task using a linear algorithm to compute from stereo images and optical flow. \cite{Valgaerts2010JointEO,Basha2010MultiviewSF} exploit energy minimization and variational methods as one the earliest works in scene flow. However, these methods are computationally expensive, which is unsuitable for applications. Other lines of work construct scene flow from a combination of optical flow and depth have demonstrated faster and good performance. \cite{Teed2021RAFT3DSF} uses rigid motion embedding to learn scene flow with an RGB-D sensors or stereo networks. \cite{Wang2020FlowNet3DGL} estimates 3D motion fields directly from a pair of point clouds. The above methods either require depth information or point clouds as prior knowledge. Since the scarcity of ground truth scene flow labels, supervised methods rely on large synthetic datasets or different input data such as RGB-D or 3D point clouds where the 3D sparse points are already presented in the data. Our focus in this paper is to show that depth and scene flow can be directly learned from unlabled images. 

Since the first introduction by \cite{Vedula2005ThreedimensionalSF}, various types of scene flow approaches have been proposed.
Early methods \cite{Valgaerts2010JointEO,Basha2010MultiviewSF} using stereo images proposed variational formulations with energy minimization and demonstrated limited accuracy and slow runtime. Recently, supervised deep learning methods directly estimate scene flow using RGB-D-image-based or point-cloud-based data and demonstrate faster run time with superior performance. 
\cite{Teed2021RAFT3DSF} uses rigid motion embedding to learn scene flow with RGB-D sensors or stereo networks.
\cite{Wang2020FlowNet3DGL} estimates 3D motion fields directly from a pair of point clouds.
Different from the works mentioned above, we focus on self-supervised monocular scene flow estimation to learn 3D points along with 3D motion field from only monocular images.

\paragraph{Monocular Depth Estimation.}
%The most recent self-supervised depth estimation methods from single images are learning based. \cite{Godard2017UnsupervisedMD} proposed a novel network architecture with training loss based on left-right consistency between images. Later, \cite{Godard2019DiggingIS} further improved monocular depth estimation via introducing minimum photometric loss and auto-masking of moving objects. \cite{Watson2019SelfSupervisedMD} investigated the issues of reprojection loss and proposed Depth Hints as a proxy label for additional supervision. \cite{Ramamonjisoa2021SingleID} used wavelet representation with deep learning to learn monocular depth map. Scarcity of ground truth depth maps lead to investigate knowledge distillation technique for self-supervised depth estimation. \cite{Peng2021ExcavatingTP} proposed a data augmentation and self-teaching loss based on distillation method to boost the performance of monocular depth estimation task. Although the mentioned methods have extensively investigated different strategies to learn better depth map, we demonstrate very competitive results compared with them using considerably simpler network without any additional supervision or tricks.  

Recent state-of-the-art self-supervised depth estimation methods from single images are learning-based. 
\cite{Godard2017UnsupervisedMD} proposed to learn monocular depth estimation from stereo pairs.
Later, \cite{Godard2019DiggingIS} proposed an improved proxy loss for handling occlusion and moving objects. \cite{Watson2019SelfSupervisedMD} introduced Depth Hints using stereo pairs as complementary supervision to boost the overall accuracy of existing self-supervised solutions.
\cite{Ramamonjisoa2021SingleID} used wavelet representation with deep learning to learn monocular depth map. 
Lack of ground truth depth maps led to investigation of knowledge distillation technique for self-supervised depth estimation. 
\cite{Peng2021ExcavatingTP} proposed a data augmentation and self-teaching loss based on the distillation method to enhance the performance of the monocular depth estimation task. 
In contrast to those methods that rely on complicated strategies, we demonstrate very competitive results by using a considerably simpler network without any additional supervision, such as pseudo labels from knowledge distillation or stereo images, or complicated training procedures.

%\paragraph{Multi-Task learning of depth and optical flow}
%Multi-task CNN-based self-supervised methods have been introduced to jointly estimate depth, optical flow, and ego-motion. Consequently, the 3D motion field can be recovered given those predicted outputs. \cite{Yin2018GeoNetUL} implicitly represents moving pixels by jointly refining depth, optical flow, and ego-motion using residual FlowNet \cite{Ilg2017FlowNet2E} without considering rigid and non-rigid objects. \cite{Zou2018DFNetUJ} exploits consistency between rigid and optical flow but only in rigid and non-occluded areas. \cite{Yang2018EveryPC} considers rigid regions and imposes consistency between depth and flow estimation without joint learning and \cite{Luo2020EveryPC} improves it by considering motion segmentation via decomposing rigid background and moving objects. \cite{Liu2019UnsupervisedLO} utilize local rigidity to fuse optical flow and depth estimation networks. Although the works mentioned earlier have achieved reasonable accuracy, they require complicated training strategies with training multiple networks in several stages. However, our method learns 3D scene flow and depth with a more straightforward training strategy and a single network in an end-to-end manner.

\paragraph{Monocular Scene Flow.}
Recent approaches to self-supervised monocular scene flow show that a 3D motion field can be directly learned from a CNN module without the need for joint training of depth, optical flow, and ego-motion. 
First, \cite{Hur2020SelfSupervisedMS} proposed to learn scene flow using a single joint decoder with proxy loss functions based on PWC-Net \cite{Sun2018PWCNetCF} architecture. 
\cite{Hur2020SelfSupervisedMS} was the first method to learn 3D motion vectors directly from the CNN module using only a sequence of images. 
Later, \cite{Hur2021SelfSupervisedMM} proposed a multi-frame monocular scene flow network to improve the previous two-view monocular scene flow model \cite{Hur2020SelfSupervisedMS} by proposing a split decoder and multi-view training strategy using convolutional LSTM. 
Despite the simplicity of the above works when compared to joint multi-task learning methods, the accuracy of these works is still lower. 
Inspired by \cite{Hur2020SelfSupervisedMS} two-view approach, we propose RAFT-MSF and substantially improve on previous two-view and multi-view models by only using two-view images as input. 
Additional to self-supervised methods, a semi-supervised method \cite{Brickwedde2019MonoSFMG} estimates scene flow from a monocular camera using probabilistic optimization framework, and \cite{Schuster2020MonoCombAS} propose a monocular combination of depth estimation and optical flow with interpolation and refinement techniques to estimate monocular scene flow. On the contrary, our fine-tuned monocular scene flow method with simpler network architecture outperforms the above semi-supervised works with faster run-time.
\section{Method}

\begin{figure*}
\centering	
\includegraphics[width=\linewidth]{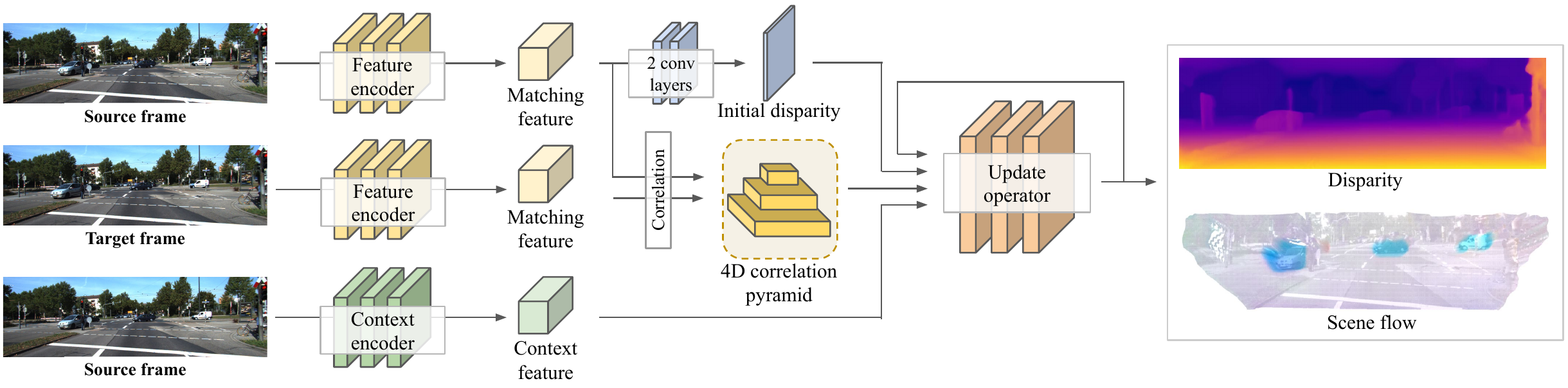}
\caption{\textbf{Overview of RAFT-MSF}. Feature networks encode input images to feature vectors. Correlation volume is constructed by using encoded features. We initialize 3D motion field to zeros everywhere and disparity is initialized using 2 convolutional layers. During each iteration, the update operator uses the current estimates of scene flow and disparity to index from the correlation pyramid. The output of the update operator is fed to scene flow and disparity heads. The scene flow can be projected to optical flow.}
\label{fig:main_fig}
\end{figure*}

Given two temporally consecutive images, $I_{t}$ and $I_{t+1}$, our model estimates a disparity map and a dense 3D motion field with respect to the camera.
Inspired by \cite{Hur2020SelfSupervisedMS,Teed2020RAFTRA}, our model iteratively estimates a 3D point \textbf{P} = $(P_{x}, P_{y}, P_{z})$ and scene flow $\textbf{s} = (s_{x}, s_{y}, s_{z})$ for each pixel \textbf{p} = $(p_{x}, p_{y})$ in the source frame, $I_{t}$.
Assuming pinhole camera model with known camera intrinsics, optical flow can be recovered by projecting scene flow and depth onto the image plane.

\subsection{Network Architecture}

% Our network architecture uses a single model to output scene flow and depth compared to other multi-task learning methods where separate networks have been utilized for each task (e.g., depth, ego-motion, and optical flow). Similar to RAFT \cite{Teed2020RAFTRA}, our network is composed of three main components: 1) a feature extractor, 2) a correlation pyramid, 3) a GRU-based update operator with scene flow and depth head modules. Our network uses the current scene flow, and depth estimates to index from the correlation volume. The GRU-based update operator takes correlation features as input to estimate the 3D motion field and depth map.

Our network consists of three main components: \textit{i)} a feature encoder, \textit{ii)} a correlation pyramid, and \textit{iii)} a GRU-based update operator with scene flow and depth head modules.
Our model first constructs a correlation pyramid using features extracted from the encoder.
Then, the GRU-based update operator takes correlation features as input and updates the 3D motion field and depth map iteratively.
The correlation features are parsed by using intermediate estimates of scene flow and depth. Here, we use the depth and disparity interchangeably; however, our model learns disparity in a hypothetical stereo setup following \cite{Godard2019DiggingIS}, the depth values can be obtained given the baseline and focal length of the camera. The overview of our RAFT-MSF is shown in \cref{fig:main_fig}.
% \todo{(better to move this paragraph after the first paragraph of Sec. 3.1.. because 'disparity' already appeared above.)}

\subsubsection{Feature Encoder}
%Given two inputs images, we first use two identical feature encoders to extract features. The first feature encoder, $f_\theta$, is used on two temporally consecutive images with shared weights to map each input image to a dense 256-dimension feature map at $1/8$ resolution. $f_\theta$ comprised of several residual blocks and downsampling layers. The second feature encoder, context encoder, initializes the hidden state of the update operator. The context encoder extracts semantic and contextual information only from the source image, $I_t$.

Given the two input images, our model extracts two kinds of features: matching feature and context feature.
For each image, the feature encoder $f_\theta$ extracts a 256-dimensional matching feature at $1/8$ resolution, which is then used for constructing a cost volume.
% The feature encoder comprises several residual blocks and downsampling layers.
The context encoder, the same architecture as in $f_\theta$ but with different trainable weights, extracts the context feature for the source image $I_t$.
The context feature provides semantic and contextual information of the source image to the update operator.

\subsubsection{Correlation Pyramid}\label{corrVolume}
%We build a full 4D correlation volume between feature vectors by employing a dot product between all pairs of feature vectors to compute visual similarity. Given encoded image features $f_\theta (I_t)\in \mathbb{R}^{HxWxD}$ and $f_\theta (I_{t+1})\in \mathbb{R}^{HxWxD}$, the correlation volume is computed as
We build a full 4D correlation volume that contains visual similarity between all possible matching pairs.
Given the per-pixel matching feature for each image, $f_\theta (I_t), f_\theta (I_{t+1})\in \mathbb{R}^{H \times W \times D}$, the correlation volume is computed as:
\begin{equation}
C_{ijkh}(I_t, I_{t+1}) = \langle f_\theta (I_t)_{ij}, f_\theta (I_{t+1})_{kh} \rangle \in\mathbb{R}^{H\times W\times H\times W},
\end{equation}
by using a dot product between matching features. $\langle \rangle$ represents a dot product operation. 
$ij$ and $kh$ refers pixel index of $I_t$ and $I_{t+1}$, respectively.

Then, we construct a 4-level correlation pyramid $\{\mathbf{C}_1, \mathbf{C}_2, \mathbf{C}_3, \mathbf{C}_4\}$ by pooling the last two dimensions of the full correlation volume,
\begin{equation}
    \mathbf{C}_k \in\mathbb{R}^{H\times W\times H/2^{k}\times W/2^{k}}.
\end{equation}
This multi-scale correlation pyramid provides information for both large and small displacements while maintaining high-resolution information.

% \todo{I think it's better to merge this paragraph with the 'inputs' paragraph under the 'update operator' section. somehow it doesn't align with the context.}
% Subsequently, we specify an indexing operator that retrieve a feature map by looking the correlation pyramid. Given a current estimate of scene flow and disparity, we map 3D motion to image plane through projection as

% \begin{equation}
%     p' = \mathbf{K}(D_{t}(p)\cdot K^{-1}p + s_{t\rightarrow t+1}(p)),
% \end{equation}
% % \todo{(notation change $x \rightarrow \textbf{p}$ )}
% with given depth map $D_t()$, scene flow $s_{t\rightarrow t+1}()$, and the camera intrinsics $\mathbf{K}$. 
% Having the current estimate of correspondence, we can index from the correlation pyramid to obtain correlation features by defining a neighborhood grid around $p'$

% \begin{equation}
%     \mathcal{N}_{p'} = \{ (p_{x} + d_{p_{x}}, p_{y} + d_{p_{y}}) | d_{p_{x}}, d_{p_{y}} \in \{-r, ..., r \} \}
% \end{equation}
% then using the local grid to sample from the correlation pyramid using bilinear sampling.
% \todo{(notation change $u, v \rightarrow p_{x}, p_{y}$ )}
 
\subsubsection{Update Operator}
Unlike RAFT that estimates optical flow, our update operator predicts a sequence of forward and backward scene flow $\{s_1, ..., s_N\}$, and disparity map of source and target frame $\{d_1, ..., d_N\}$. The bi-directional \cite{Meister2018UnFlowUL} estimation allows to utilize occlusion cues. For simplicity, we only show the forward scene flow and the disparity of source frame.
At each iteration, the update operator residually updates the estimates,
\begin{equation}
    s_{k+1} = \Delta s + s_{k}, d_{k+1} = \Delta d + d_{k}.
    \label{eq:residual}
\end{equation}

\begin{figure}
\centering	
\includegraphics[width=\linewidth]{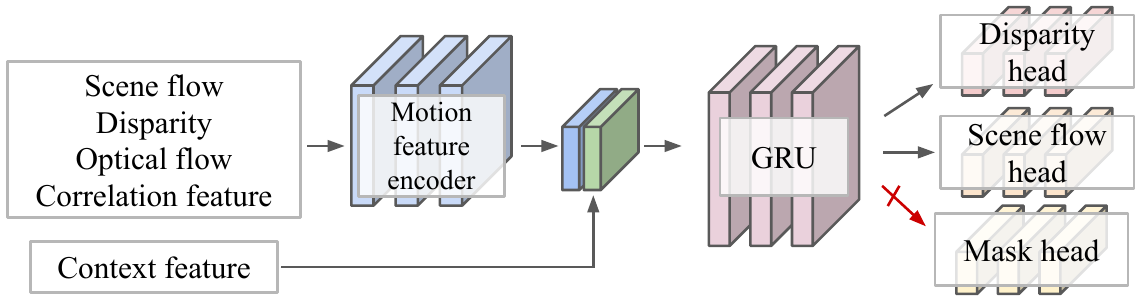}
\caption{\textbf{A detailed view of the update operator}: The motion feature encoder encodes previous estimates disparity, scene flow, optical flow, and correlation features to motion features, which is concatenated with context features and passed to GRU. Disparity and scene flow heads are used to decode disparity and scene flow estimates, and mask head is used to produce a convex mask for upsampling. }
\label{fig:update_fig}
\end{figure}

\cref{fig:update_fig} illustrates an overview of the update operator.
The update operator is based on a GRU and takes features from the correlation pyramid via indexing operator, current estimates of scene flow and disparity, and a hidden state from the previous iteration step, then outputs the residual update $\Delta s$ and $\Delta d$ for scene flow and disparity, respectively. 
It additionally outputs an upsampling mask from the mask head.
The input hidden state of the update operator is initialized by the context encoder with the $tanh$ function as activation.

% update operator figure
% \input{figures/update_fig}

\paragraph{Initialization.} We initialize scene flow to zeros (\ie, $s_{0} = 0$) but disparities with convolutional layers.
We found that initializing disparities with two convolutional layers yields better accuracy for both disparity and scene flow compared to initializing with zeros.
Given the feature map of source image, we initialize our disparity:
\begin{equation}
    d_{0} = g_w(f_\theta(I_t)),
\end{equation}
with two convolutional layers $g_w()$.
% \todo{(probably better to visualized the learned disparity map? and justify, for example..: though it doesn't explicitly demonstrate meaningful context, it improves the accuracy with minimal computational cost.)}.

\paragraph{Inputs.} 
At each iteration step, we retrieve correlation features for the residual updates from the pre-computed 4D correlation pyramid $\mathbf{C}_k$.
Given current estimates scene flow $s_k$, disparity $d_k$, and the camera intrinsics $\mathbf{K}$, we first calculate the corresponding pixel $\textbf{p}'$ for each pixel \textbf{p},
\begin{equation}
    \textbf{p}' = \mathbf{K}(d_{t}(\textbf{p})\cdot K^{-1}\textbf{p} + s_{t\rightarrow t+1}(\textbf{p})),
\end{equation}
and retrieve correlation features of a set of pixels $\mathcal{N}_{\textbf{p}'}$ neighboring the corresponding pixel $\textbf{p}'$, within a range of $[-r, r]$,
\begin{equation}
    \mathcal{N}_{\textbf{p}'} = \{ (p_{x} + d_{p_{x}}, p_{y} + d_{p_{y}}) | d_{p_{x}}, d_{p_{y}} \in \{-r, ..., r \} \}.
\end{equation}
Of course, optical flow can be naturally obtained by $\mathbf{F} = \textbf{p}' - \textbf{p}$.
Given the current estimates and the retrieved correlation feature, we pass them through the motion feature encoder to get a motion feature.
Then the GRU takes a concatenation of the motion features and the context features (from the context encoder) as an input.

% We specify a lookup operator $\mathbf{L_{C}}$ that produces a correlation feature map by indexing the correlation pyramid. Given a current estimate of scene flow $s_k$, disparity $d_k$, and the camera intrinsics $\mathbf{K}$, we map 3D motion to image plane through projection as

% \begin{equation}
%     p' = \mathbf{K}(d_{t}(p)\cdot K^{-1}p + s_{t\rightarrow t+1}(p)),
% \end{equation}
% % \todo{(notation change $x \rightarrow \textbf{p}$ )}
% Having the current estimate of correspondence $p' = (p_{x}, p_{y})$, we can index from the correlation pyramid to obtain correlation features by defining a neighborhood grid around $p'$ as
% \begin{equation}
%     \mathcal{N}_{p'} = \{ (p_{x} + d_{p_{x}}, p_{y} + d_{p_{y}}) | d_{p_{x}}, d_{p_{y}} \in \{-r, ..., r \} \}
% \end{equation}
% then using the local grid to sample from the correlation pyramid using bilinear sampling. Hence, we can recover 2D flow field as $\mathbf{F = p' - p}$.

% To process the input features, we employ two convolutional layers for indexed correlation features $L_{C}(p')$, the 2D flow field $\mathbf{F}$, and the current estimates of scene flow and disparity separately to generate 2D motion features. The output motion features from the motion feature encoder is the concatenation of processed input features, see \cref{fig:update_fig}. Finally, the GRU update operator takes the concatenation of the encoded 2D motion features with the input from the context network. An overview of update operator is given in \cref{fig:update_fig}.

\paragraph{Predictions.} The crucial part of the update operator is GRU units with convolution layers instead of fully connected layers. The encoded motion features are iteratively decoded to predict residual scene flow and disparity. 
As illustrated in \cref{fig:update_fig}, the output of the GRU is fed to scene flow and disparity heads to estimate corresponding outputs. We employ three convolutional layers to predict scene flow and disparity residuals, respectively. 
The final prediction is the sum of all residual outputs with initial values (\rf~\cref{eq:residual}).

\begin{figure}[t]
    \centering
    \subcaptionbox{Input}{\includegraphics[width=0.8\linewidth]{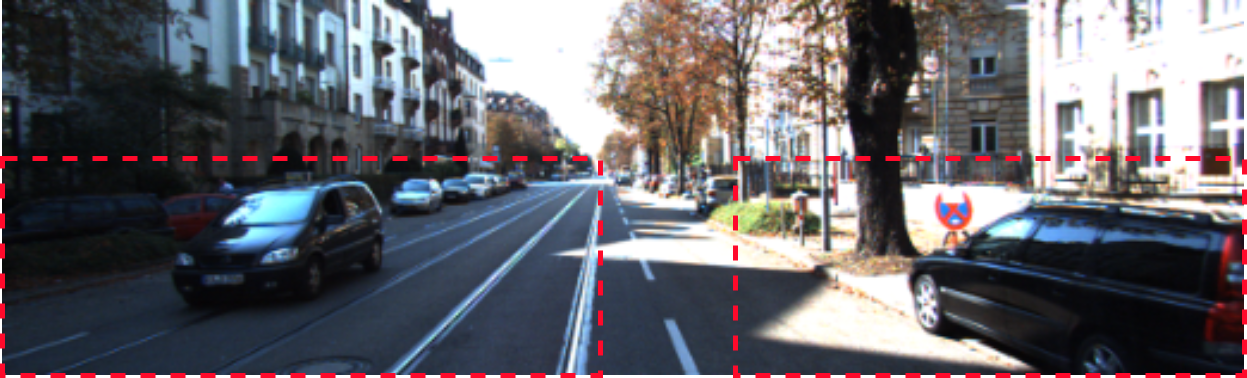}}
    \subcaptionbox{Without detach}{\includegraphics[width=0.8\linewidth]{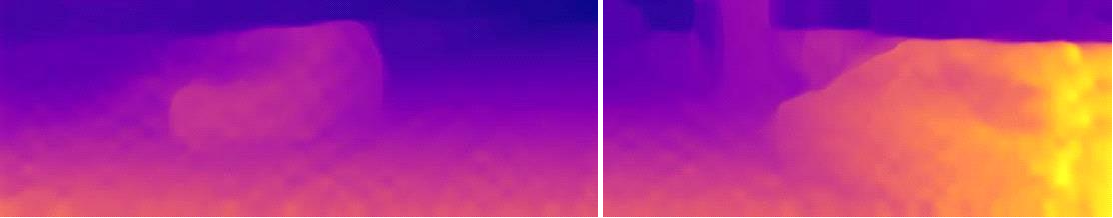}}
    \subcaptionbox{Detach}{\includegraphics[width=0.8\linewidth]{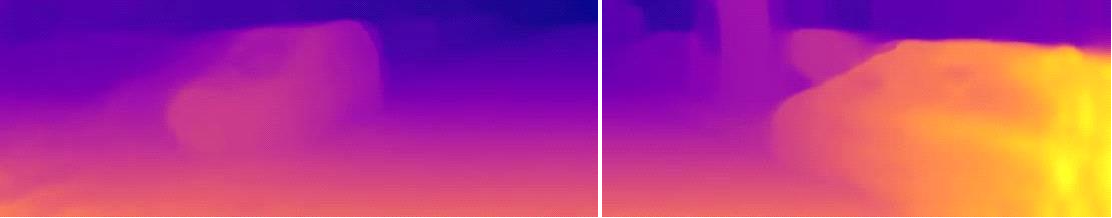}}
    \caption{\textbf{Ablation study of our detaching technique}. Dashed lines in (a) source frame is cropped and visualized in (b) without detaching and (c) with detaching the weights of the output of the GRU. More qualitative results are available in the supplementary material.}
    \label{fig:detach_vis_abl}
\end{figure}

\paragraph{Upsampling Module.} 
The predicted disparity and scene flow estimates are at 1/8 of the input resolution.
Then a learning-based upsampling module using convex upsampling \cite{Teed2020RAFTRA} to upsample the output to the input image resolution.
The upsampling module takes the feature map from the output of the GRU and processes it using two convolutional (\cref{fig:update_fig}, mask head) layers to generate a convex mask. 
However, we found that in the self-supervised learning setup, direct usage of the convex upsampling operator results in checkerboard artifacts on predicted outputs at later stages (see \cref{tab:testBenc}), and the model fails to converge.
To address it, we detach the gradients of the hidden state outputted by the GRU in each upsampling operation. 
In \cref{fig:update_fig}, the arrow \textcolor{red}{$\not\to$} represents detaching the output of GRU for the mask head at each iteration.
In the ablation study in \cref{tab:abl1}, we demonstrate that the gradient detaching technique further improves the accuracy. Moreover, in \cref{fig:detach_vis_abl}, using detaching technique results in a smooth and artifact-free depth map while without detaching, we get distorted and checkerboard artifacts on depth map.

\subsection{Self-supervised learning}

Similar to previous methods, we also use view synthesis as a proxy task for learning depth and scene flow in a self-supervised manner.
Given depth and scene flow estimates, we synthesize the reconstructed source image $\hat{I}_{t}$ from the target frame $I_{t+1}$ and penalize the photometric difference to the source image $I_{t}$ so that the network outputs a correct combination of depth and scene flow.

% Photometric-error-based on brightness consistency assumption is the major learning objective for monocular reconstruction methods and it evaluates the photometric similarity between source image $I_{t}$ and reconstructed image $\hat{I}_{t}$ through view synthesis. In our case, given depth and scene flow estimates of the source image, we can synthesize $\hat{I}_{t}$ via the target frame, . 

We build our loss functions following \cite{Hur2020SelfSupervisedMS},
\begin{equation}\label{total_loss}
\begin{split}
    \mathcal{L}_{total} = \mathcal{L}_{d\_ph} + \lambda_{d\_sm}\mathcal{L}_{d\_sm} + \mathcal{L}_{sf\_ph} \\
    + \lambda_{sf\_sm}\mathcal{L}_{sf\_sm} + \lambda_{sf\_pnt}\mathcal{L}_{sf\_pnt},
\end{split}
\end{equation}
which is a weighted combination of photometric consistency $\mathcal{L}_{ph}$, edge-aware smoothness $\mathcal{L}_{sm}$, and 3D point reconstruction loss $\mathcal{L}_{pnt}$ for scene flow and disparity estimates.
We also uses stereo pairs for training, but in the test time, of course the input is purely monocular images.
More details can be found in \cite{Hur2020SelfSupervisedMS}.

Given $n$ intermediate predictions, we apply the self-supervised loss on the entire sequence of predictions.
We exponentially decay the weights of each loss in \cref{total_loss},
\begin{equation}\label{seqLoss}
    \mathcal{L}_\text{sequence} = \sum_{i=1}^{n}\gamma^{n-i}\mathcal{L}_{i},
\end{equation}
where $n$ is the number of disparity and scene flow iterations, $\gamma$ is the decay factor, and $\mathcal{L}_{i}$ is the loss at each $i^{th}$ iteration step.

% It is a common approach to apply losses on intermediate predictions of supervised models. However, this is not an optimal fit for un/self-supervised learning methods because intermediate outputs are at lower resolution. \cite{Hur2020SelfSupervisedMS} upsample all intermediate outputs to match with ground truth resolution and then calculate losses. Different from \cite{Hur2020SelfSupervisedMS}, our model's intermediate results are at high resolution; thus we apply the self-supervised loss on entire sequence of predictions without losing information. We exponentially decay the weights of each loss in \cref{total_loss},
% \begin{equation}\label{seqLoss}
%     \mathcal{L}_{sequence} = \sum_{i=1}^{n}\gamma^{n-i}\mathcal{L}_{i},
% \end{equation}
% where $n$ is the number of disparity and scene flow iterations, $\gamma$ is the decay factor, and $\mathcal{L}_{i}$ is the loss at each $i^{th}$ iteration step.

\section{Experiments}

%Extensive experiments are carried out to benchmark RAFT-SF against state-of-the-art scene flow methods. To keep consistency with previous works \cite{Hur2020SelfSupervisedMS,Hur2021SelfSupervisedMM}, we use the same dataset, KITTI raw \cite{Geiger2013VisionMR}, and same protocol for all experiments. Particularly, we use KITTI split \cite{Godard2017UnsupervisedMD} without including 29 scenes that are present in KITTI Scene Flow Training \cite{Menze2015ObjectSF}. Afterwards, we split the remaining 32 scenes into \num{25801} sequence of images of training and \num{1684} images for validation samples. We use KITTI Scene Flow Training to evaluate and ablate our model, since it has ground truth labels for scene flow and disparity for 200 images. 

For a fair comparison with the state-of-the-art methods, we follow the same training dataset (KITTI raw \cite{Geiger2013VisionMR}) and protocols from our direct previous work \cite{Hur2020SelfSupervisedMS,Hur2021SelfSupervisedMM}.
For the main experiment and ablation study, we use KITTI split \cite{Godard2017UnsupervisedMD} consisting of \num{25801} training pairs and \num{1684} validation pairs.
Then, we test our model on the KITTI Scene Flow Training, which contains scene flow and disparity ground truth for 200 image pairs.
For monocular depth evaluation, we additionally use the Eigen Split \cite{Eigen2014DepthMP} that consists of \num{20120} training pairs and 1338 validation pairs.
Note that there is no overlap between the training data and testing data. Following \cite{Hur2020SelfSupervisedMS}, we also perform fine-tuning of our monocular scene flow model.

\subsection{Implementation Details}
For training, we use Adam optimizer \cite{Kingma2015AdamAM} ($\beta_{1} = 0.9$, $\beta_{2} = 0.999$) and clip gradients to the range of $[-1, 1]$ with initial learning rate of \num{0.0001}.
The training takes about two days on 4 GPUs.
We train our model for \num{200}k iterations with the mini-batch size of 8. 
Our method does not require a complicated stage-wise pre-training; thus, it can be trained at once. 
Unless otherwise noted, we set the number of update iteration $n$ in \cref{seqLoss} to 10 and the weight decay factor $\gamma$ to 0.8. 
More details of the hyperparameters in the self-supervised loss (\cref{total_loss}) can be found in \cite{Hur2020SelfSupervisedMS}.

\begin{table}[!htb]
\centering
\footnotesize
\setlength\tabcolsep{3pt}
\begin{tabular*}{\columnwidth}{@{\extracolsep{\fill}}cc@{\hskip 4em}S[table-format=2.2]S[table-format=2.2]S[table-format=2.2]S[table-format=2.2]@{}}
	\toprule
	Grad. detach & Disp. init. & {D$1$-all} & {D$2$-all} & {Fl-all} & {SF-all} \\
	\midrule
	\multicolumn{2}{l}{\scriptsize \quad \quad \quad \emph{(RAFT baseline)}} & 26.95  & 28.11  & 18.01  & 38.37   \\
	\checkmark &        					& 25.48  & 27.73  & 17.26  & 37.37   \\
	\checkmark & \checkmark & \bfseries 24.04  & \bfseries 27.06  & \bfseries 16.62  & \bfseries 35.62  \\
	\bottomrule
\end{tabular*}
\caption{\textbf{Ablation study on our contributions}: We first propose a monocular scene flow RAFT baseline (\emph{RAFT baseline}) which already outperforms the previous state of the arts. The gradient detaching on the upsampling module and disparity initialization with convolutional layers further improve the accuracy.}
\label{tab:abl1}
\end{table}

\subsection{Ablation Study}

% qualitative figure
\begin{figure*}[!htb]
\centering
\scriptsize
\setlength\tabcolsep{0.1pt}
\renewcommand{\arraystretch}{0.2}
\begin{tabular}{>{\centering\arraybackslash}m{.03\textwidth} >{\centering\arraybackslash}m{.23\textwidth} >{\centering\arraybackslash}m{.23\textwidth} @{\enspace} >{\centering\arraybackslash}m{.23\textwidth} >{\centering\arraybackslash}m{.23\textwidth}}
% \multirow{4}{*}{\rotatebox[origin=c]{90}{KITTI}}
    & Source frame & Target frame & Source frame & Target frame   \\[0.5em]
    \rotatebox[origin=c]{90}{Input} &
	\includegraphics[width=\linewidth]{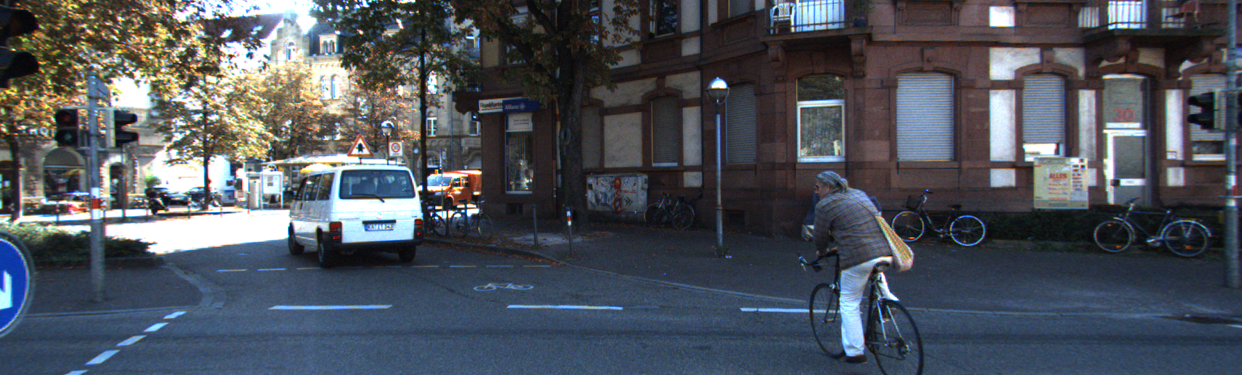} & 
	\includegraphics[width=\linewidth]{figures/00000210.png} &
	\includegraphics[width=\linewidth]{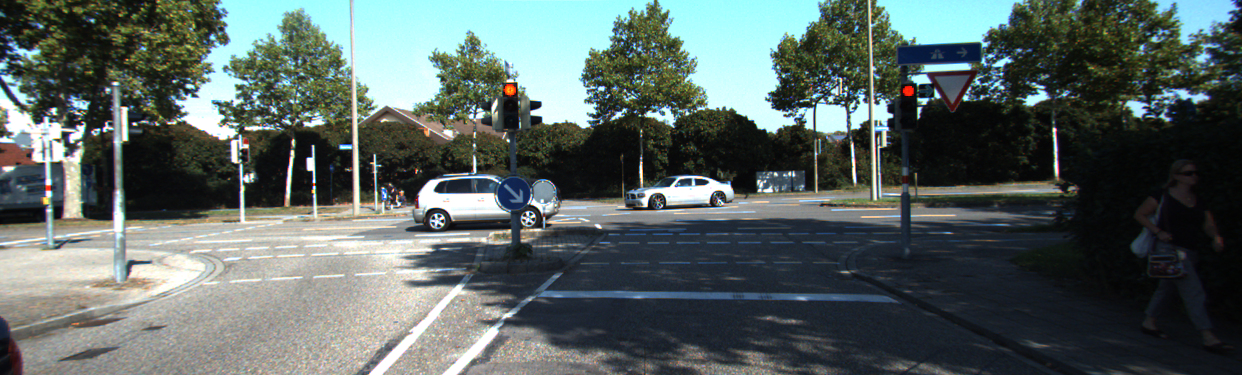} &
	\includegraphics[width=\linewidth]{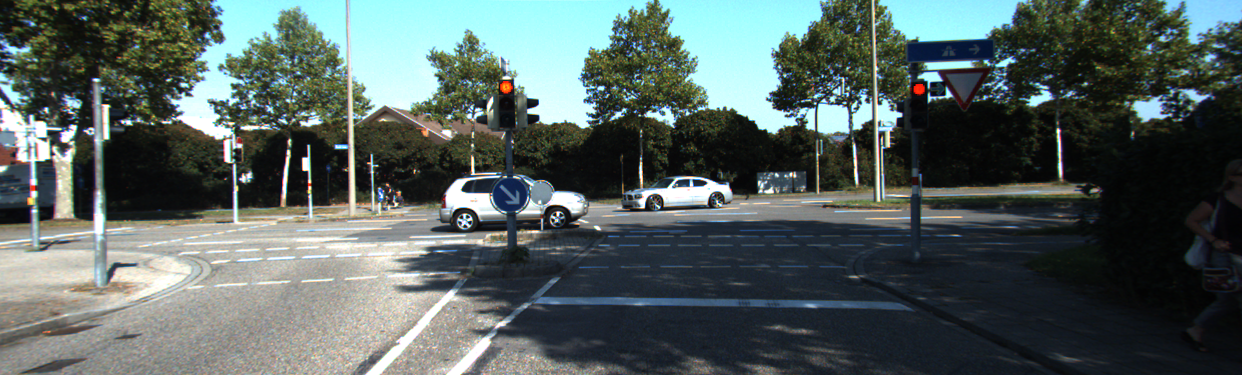} \\ [3em] \\
	
	& Disparity & Disparity error map & Disparity & Disparity error map \\
    \rotatebox[origin=l]{90}{\textbf{Ours}} &
    \includegraphics[width=\linewidth]{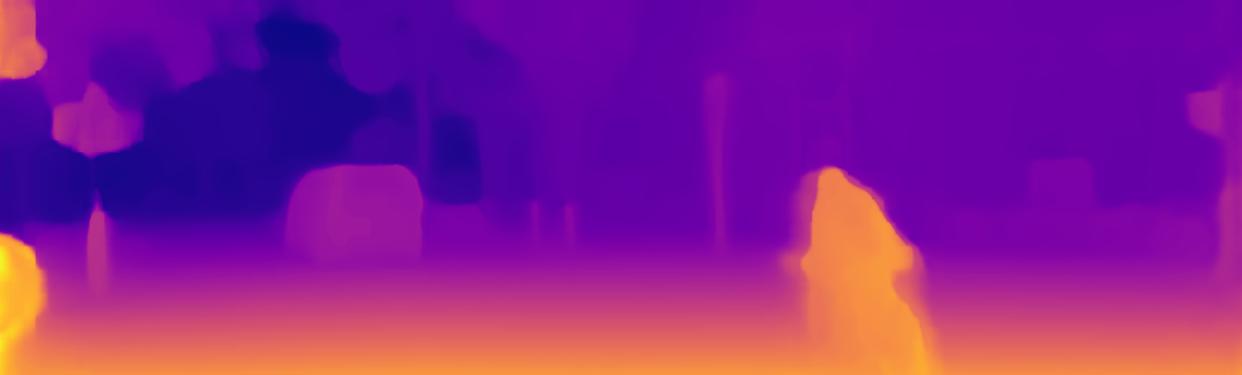} &
	\includegraphics[width=\linewidth]{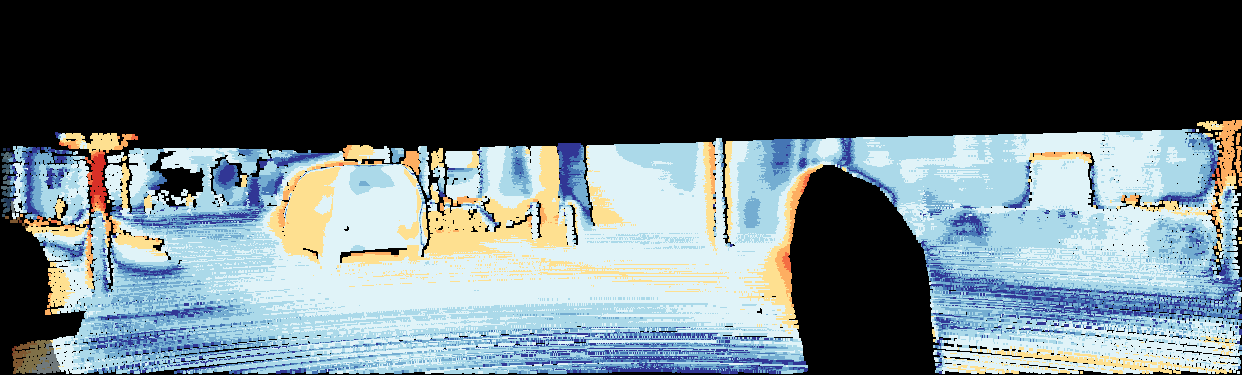} &
	\includegraphics[width=\linewidth]{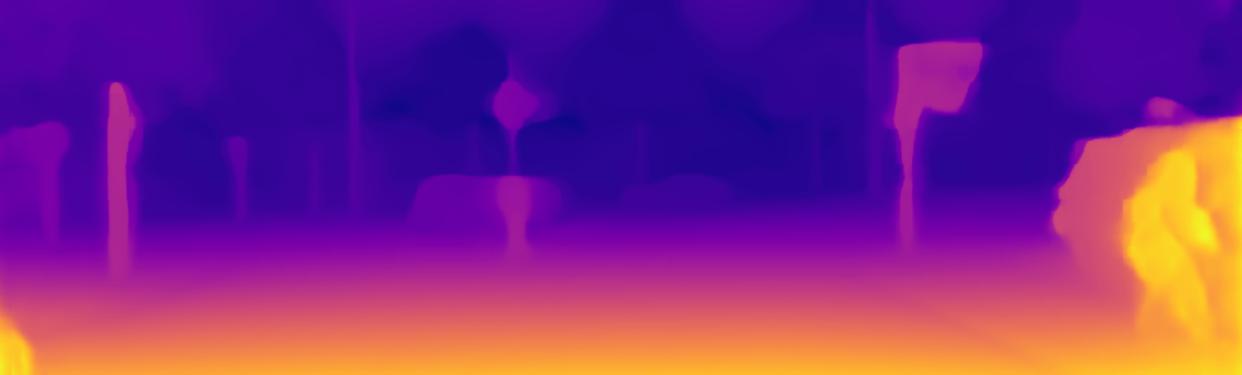} &
	\includegraphics[width=\linewidth]{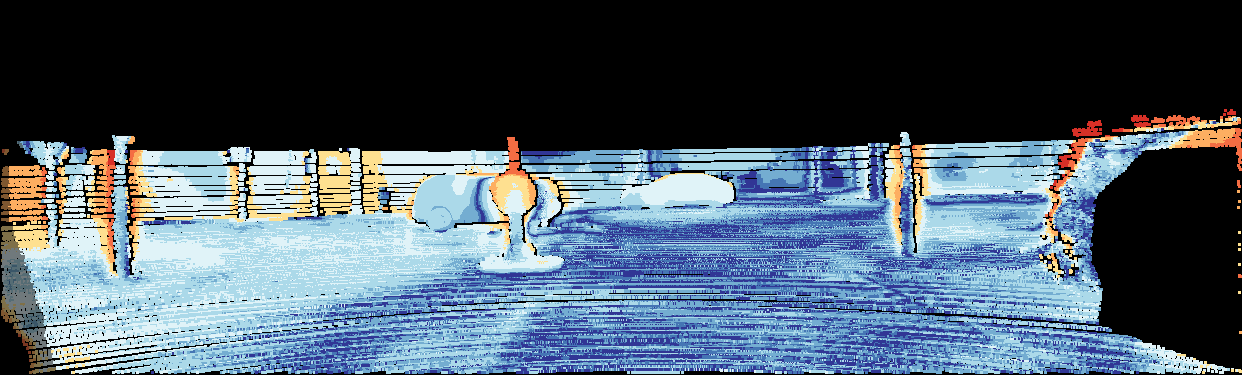} \\ [0.5em] \\
	
	\tiny \rotatebox[origin=l]{90}{Self-Mono-SF} &
	\includegraphics[width=\linewidth]{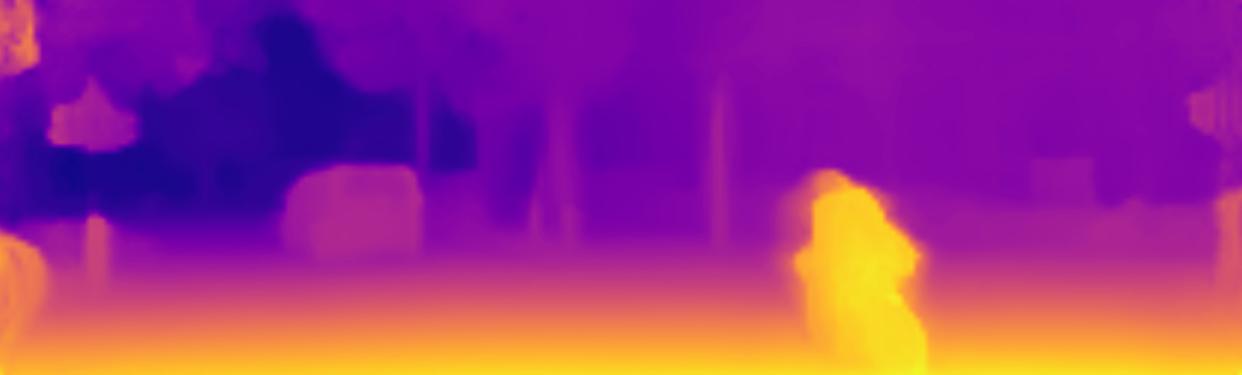} &
	\includegraphics[width=\linewidth]{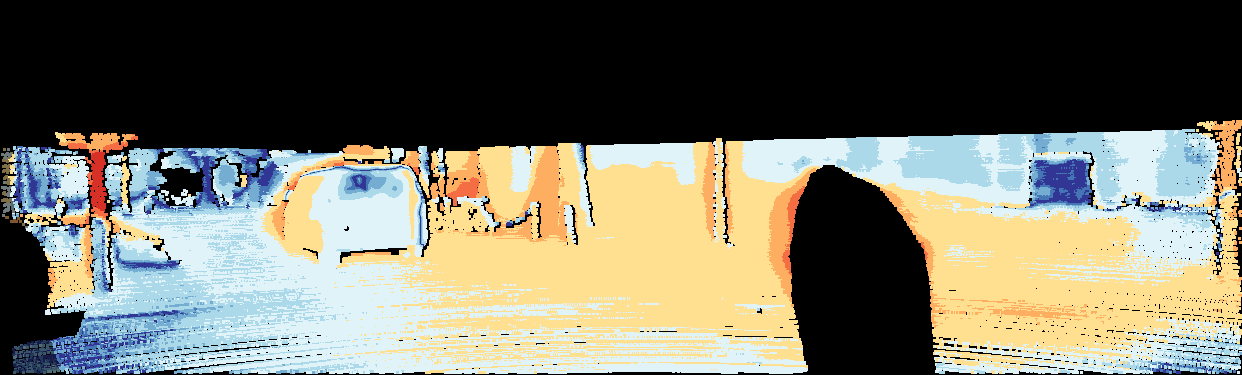} &
	\includegraphics[width=\linewidth]{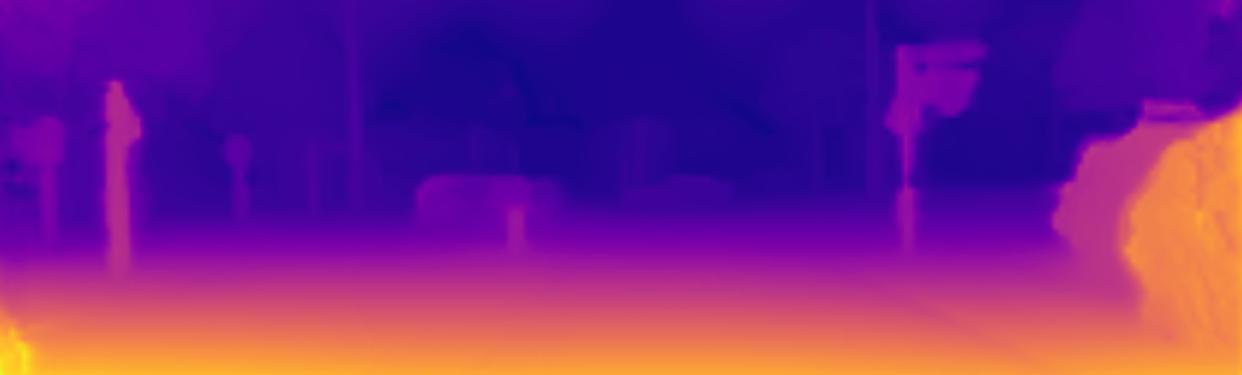} &
	\includegraphics[width=\linewidth]{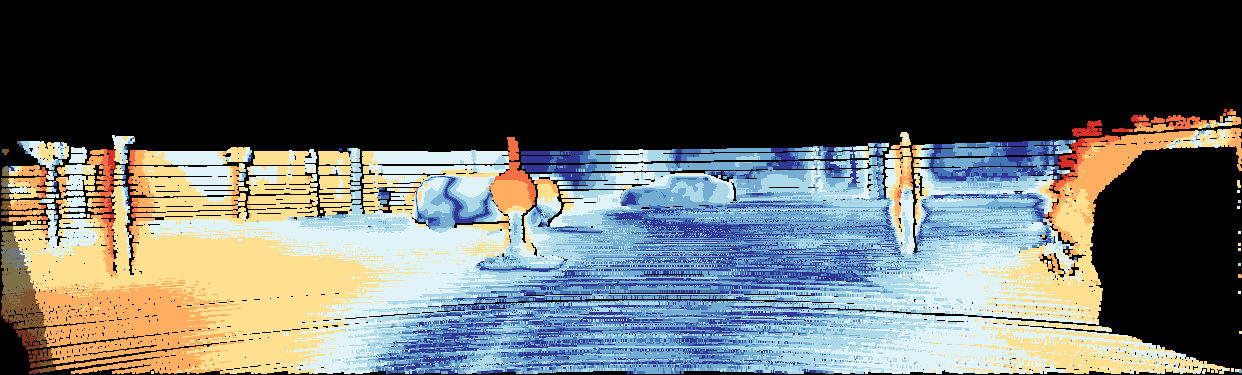} \\ [3em] \\
	
	& Scene flow & Scene flow error map & Scene flow & Scene flow error map \\
	\rotatebox[origin=l]{90}{\textbf{Ours}} &
	\includegraphics[width=\linewidth]{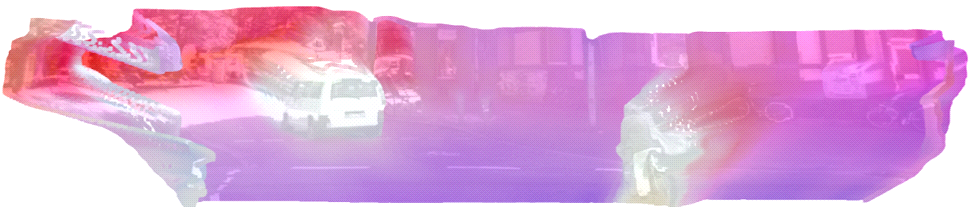} &
	\includegraphics[width=\linewidth]{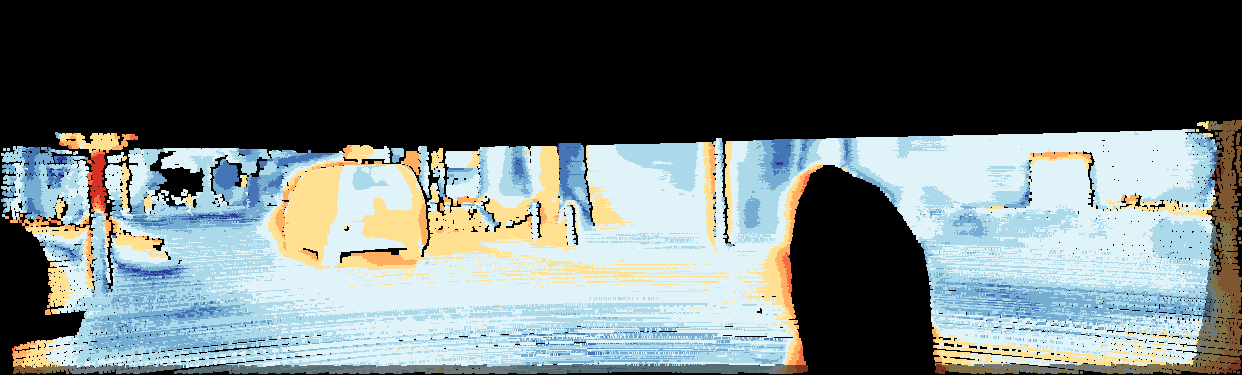} &
	\includegraphics[width=\linewidth]{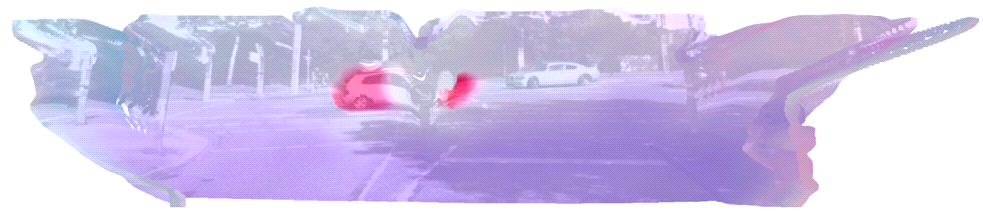} &
	\includegraphics[width=\linewidth]{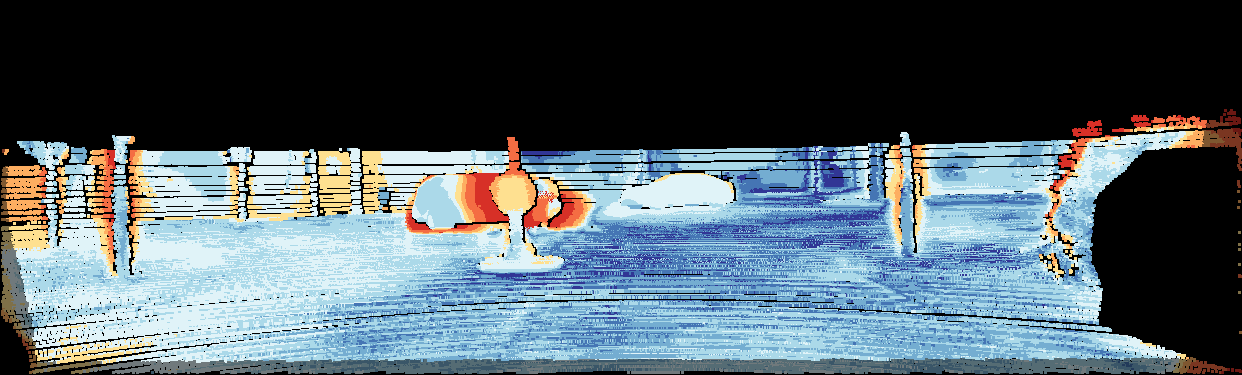} \\ [0.5em] \\
	
	\tiny \rotatebox[origin=l]{90}{Self-Mono-SF} &
	\includegraphics[width=\linewidth]{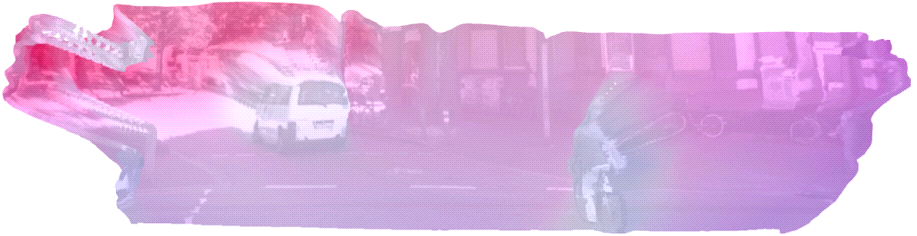} &
	\includegraphics[width=\linewidth]{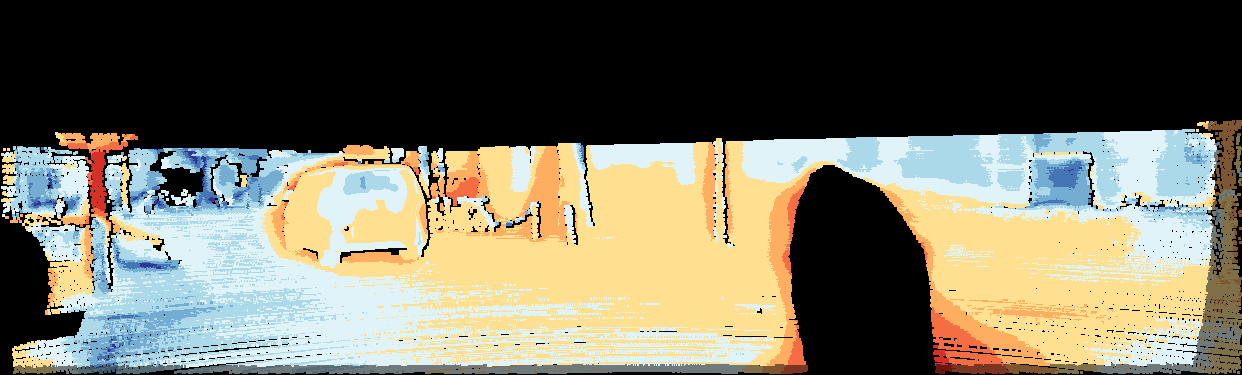} &
	\includegraphics[width=\linewidth]{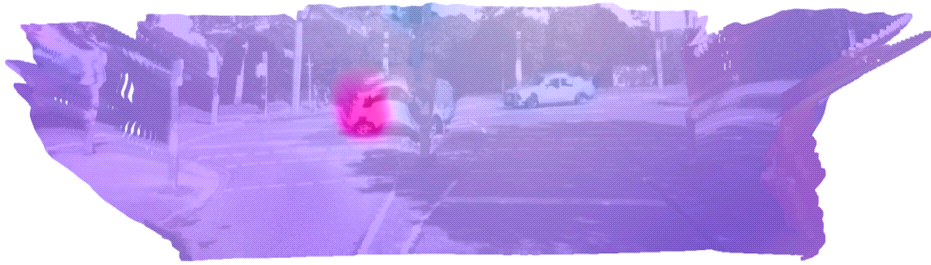} &
	\includegraphics[width=\linewidth]{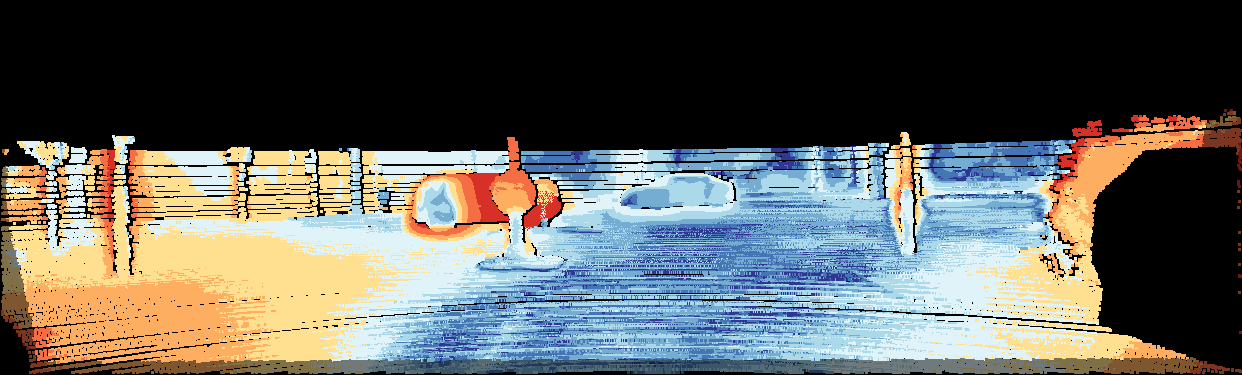} \\ 
\end{tabular}
\caption{\textbf{Qualitative comparison of our model (RAFT-MSF) with Self-Mono-SF \protect\cite{Hur2020SelfSupervisedMS}}: top row shows two temporal input images ($I_{t}, I_{t+1}$), second and third rows compare depth and depth error maps of our method (RAFT-MSF) and Self-Mono-SF \protect\cite{Hur2020SelfSupervisedMS}, and last two rows compare scene flow and scene flow error maps of the same methods, respectively. More qualitative results can be found in supplementary materials.}
\label{fig:image}
\end{figure*}

To validate our contributions, we conduct ablation studies on our model, RAFT-MSF. 
For the ablations studies, we set the number of update iterations $n$ to 6 for fast convergence.
We use the scene flow metrics for evaluation, which is defined as the outlier rate in $\%$. 
A pixel is considered an outlier if it exceeds a threshold of 3 pixels and 5$\%$ w.r.t. the ground truth labels. 
The scene flow outlier rate (SF-all) is obtained after calculating the outlier rate of the disparity (D1-all), disparity change (D2-all), and optical flow (Fl-all).

\Cref{tab:abl1} demonstrates the ablation study of our major contributions.
Given the vanilla RAFT optical flow backbone, we demonstrate a self-supervised monocular scene flow pipeline on RAFT \emph{(RAFT baseline)} in the first row.
This first contribution already outperforms the previous best two-frame self-supervised method \cite{Hur2020SelfSupervisedMS} by a large margin (38.27 \% \vs~47.05 \%).

Our contributions using gradient detach for convex upsampling layers and disparity initialization further improve the accuracy up to 7.2\% from our monocular scene flow proposal on RAFT.
Especially, the gradient detaching strategy effectively removes the checkerboard artifacts appearing during the upsampling process and outputs clear object boundaries. Specifically, in \cref{fig:detach_vis_abl} (b), the cropped cars and the ground are checkered and distorted; however, in \cref{fig:detach_vis_abl}(c), the visualized depth maps are smoother and checker free.

In \cref{tab:abl2}, we further investigate the different types of disparity initialization.
Comparing to simply initializing with zero, using convolutional layers for initialization yields better accuracy.
We choose to use two convolutional layers for better accuracy with the minimal burden of computational cost.
We also tried different initialization types, filling with one or a half of the maximum disparity, but the training did not converge.

\begin{table}[t]
\centering
\footnotesize
\begin{tabularx}{\columnwidth}{@{} X  @{\hskip 4.3em} S[table-format=2.2] @{\hskip 1em} S[table-format=2.2] @{\hskip 1em} S[table-format=2.2] @{\hskip 1em} S[table-format=2.2] @{}}
	\toprule
	Type of dispairty init.   & {D1-all} & {D2-all} & {Fl-all} & {SF-all} \\
	\midrule
	Zero  & 25.48  & 27.73  & 17.26  & 37.37   \\
    1 convolutional layer   & 24.52  & 27.40  & 16.90  & 36.15   \\
    2 convolutional layers  & \bfseries 24.04  & \bfseries 27.06  & \bfseries 16.62  & \bfseries 35.62  \\
	\bottomrule
\end{tabularx}
\caption{\textbf{Ablation study on disparity initialization:} Initializing disparity with convolutional layers outperforms initializing with zeros. We choose to use two layers which give better accuracy without much computational cost.}
\label{tab:abl2}
\end{table}

% mono_depth table
\begin{table*}[!t]
\centering
\scriptsize
\begin{tabularx}{0.8\linewidth}{@{}c@{\hskip 0.7em}XS[table-format=1.3,round-precision=3]@{\hskip 0.6em}S[table-format=1.3,round-precision=3]@{\hskip 0.6em}S[table-format=1.3,round-precision=3]@{\hskip 0.6em}S[table-format=1.3,round-precision=3]S[table-format=1.3,round-precision=3]@{\hskip 0.6em}S[table-format=1.3,round-precision=3]@{\hskip 0.6em}S[table-format=1.3,round-precision=3]@{}}
	\toprule
	 
	{Split} & {Method} & {Abs Rel $\downarrow$} & {Sq Rel $\downarrow$} & {RMSE $\downarrow$}  & {RMSE log $\downarrow$} & {$\delta < 1.25$ $\uparrow$}    & {$\delta < 1.25^2$ $\uparrow$}    & {$\delta < 1.25^3$ $\uparrow$}    \\ \midrule 
	\multirow{4}{*}{\rotatebox[origin=c]{90}{KITTI}}
	  & \cite{Yang2018EveryPC}          & 0.109   & 1.004  & 6.232 & 0.203    & 0.853 & 0.937 & 0.975 \\
      & \cite{Liu2019UnsupervisedLO}          & 0.108   & 1.020  & 5.528 & 0.195    & 0.863 & 0.948 & 0.980 \\
      & \cite{Hur2020SelfSupervisedMS} & 0.106   & 0.888  & 4.853 & 0.175    & 0.879 & 0.965 & \bfseries 0.987 \\
      & \textbf{RAFT-MSF (ours)}         & \bfseries 0.082  & \bfseries 0.726  & \bfseries 4.165 & \bfseries 0.148    & \bfseries 0.921 & \bfseries 0.971 & 0.986 \\
	\midrule
	\multirow{9}{*}{\rotatebox[origin=c]{90}{Eigen}}
      & \cite{Ranjan2019CompetitiveCJ}             & 0.155   & 1.296  & 5.857 & 0.233    & 0.793 & 0.931 & 0.973 \\
      & \cite{Chen2019SelfSupervisedLW}          & 0.135   & 1.070  & 5.230 & 0.210    & 0.841 & 0.948 & 0.980 \\
      & \cite{Yang2018EveryPC}            & 0.127   & 1.239  & 6.247 & 0.214    & 0.847 & 0.926 & 0.969 \\
      & \cite{Luo2020EveryPC}          & 0.127   & 0.936  & 5.008 & 0.209    & 0.841 & 0.946 & 0.979 \\
      & \cite{Hur2020SelfSupervisedMS}   & 0.125   & 0.978  & 4.877 & 0.208    & 0.851 & 0.950 & 0.978 \\
      & \cite{Godard2019DiggingIS}           & 0.105   & 0.822  & 4.692 & 0.199    & 0.876 & 0.954 & 0.977 \\
      & \cite{Watson2019SelfSupervisedMD}         & 0.099   & 0.723  & 4.445 & 0.187    & 0.886 & 0.961 & 0.982 \\
      & \cite{Ramamonjisoa2021SingleID}         & 0.102   & 0.739  & 4.452 & 0.188    & 0.883 & 0.960 & 0.981 \\
      & \cite{Zhou_2021_ICCV}         & 0.112   & 0.753  & 4.530 & 0.189    & 0.881 & 0.961 & 0.982 \\
      & \cite{Peng2021ExcavatingTP}         & \bfseries 0.093   & \bfseries 0.671  & \bfseries 4.297 & \bfseries 0.178    & 0.899 & \bfseries 0.965 & \bfseries 0.983 \\
      & \textbf{RAFT-MSF (ours)}           & \bfseries 0.093  & 0.781  & 4.321 & 0.186    & \bfseries 0.901 & 0.960 & 0.981 \\
	\bottomrule
\end{tabularx}
\caption{\textbf{Monocular depth comparison}: We show a superior accuracy on KITTI Split compared to published multi-task methods. In Eigen split, our method outperforms multi-task methods and deliver competitive results with single monocular depth estimation tasks. ($\downarrow$: lower is better, and $\uparrow$: higher is better)}
\label{tab:depthTab}
\end{table*}

% kitti scene flow train table
\begin{table}[t]
\centering
\footnotesize
\begin{tabularx}{\columnwidth}{@{}XS[table-format=2.2]@{\hskip 0.7em}S[table-format=2.2]@{\hskip 0.7em}S[table-format=2.2]@{\hskip 0.6em}S[table-format=2.2]@{}}
	\toprule
	Method    &   {D$1$-all}   &  {D$2$-all}  & {Fl-all}   & {SF-all}  \\
	\midrule
	\cite{Zou2018DFNetUJ}        & 46.50  & 61.54  & 27.47  & 73.30 \\
    \cite{Yin2018GeoNetUL}        & 49.54  & 58.17  & 37.83  & 71.32 \\
    \cite{Luo2020EveryPC}         & 23.84  & 60.32  & 19.64  & {(\textgreater{}60.32)} \\
    \cite{Hur2020SelfSupervisedMS}  & 31.25  & 34.86  & 23.49  & 47.05 \\
    \cite{Hur2021SelfSupervisedMM} & 27.33  & 30.44  & 18.92  & 39.82  \\
    \textbf{RAFT-MSF (ours)} & \bfseries 18.34  & \bfseries 23.65 & \bfseries 17.51  & \bfseries 30.97 \\

	\bottomrule
\end{tabularx}
\caption{Monocular scene flow evaluation on \textbf{KITTI 2015 Scene Flow Training}: Our self-supervised method significantly outperforms both multi-task CNN methods and recently published monocular scene flow methods on all the metrics.}
\label{tab:kitti2015Train}
\end{table}

\subsection{Monocular Scene Flow}

We compare our full model (RAFT-MSF) on the KITTI Scene Flow Training and Test benchmark. \Cref{tab:kitti2015Train} shows the accuracy comparison with state-of-the-art self-supervised monocular scene flow methods on KITI Scene Flow Training set.
Our method achieves the best scene flow accuracy among self-supervised multi-task CNN methods and self-supervised monocular scene flow methods by a large margin. 
Especially, our model outperforms the best two-frame self-supervised method \cite{Hur2020SelfSupervisedMS} by $34.17\%$ and even multi-frame self-supervised method \cite{Hur2021SelfSupervisedMM} by $22.23\%$.

On KITTI Scene Flow 2015 Test benchmark in \cref{tab:testBenc}, our self-supervised model consistently outperforms the state of the arts, reducing the scene flow error by $29.39\%$ on two-view \cite{Hur2020SelfSupervisedMS} and $20.57\%$ on multi-view method \cite{Hur2021SelfSupervisedMM}, respectively.
We also fine-tune our self-supervised model using 200 pairs of ground truth in a semi-supervised manner and test on KITTI Scene Flow 2015 benchmark.
Our fine-tuned model (RAFT-MSF-ft) also achieves the best scene flow accuracy among all published semi-supervised methods, especially outperforming the best monocular method \cite{Brickwedde2019MonoSFMG} with 228 times faster runtime (0.18 (s) \vs~ 41 (s)).

Qualitative results in \cref{fig:image} show the superiority of our method over \cite{Hur2020SelfSupervisedMS}, improving the disparity and scene flow estimation on planar road surface and object boundaries.

\subsection{Monocular depth}

% kitti scene flow test table
\begin{table}[t]
\centering
\footnotesize
\begin{tabularx}{\columnwidth}{@{}XS[table-format=2.2]@{\hskip 0.7em}S[table-format=2.2]@{\hskip 0.7em}S[table-format=2.2]@{\hskip 0.6em}S[table-format=2.2]@{}}
	\toprule
	Method    &   {D$1$-all}   &  {D$2$-all}  & {Fl-all}   & {SF-all}  \\
	\midrule
\cite{Brickwedde2019MonoSFMG}           & 16.32             & 19.59             & 12.77          & 23.08           \\
\cite{Yang2020UpgradingOF}              & 25.36             & 28.34             & \bfseries 6.30 & 30.96           \\
\cite{Schuster2020MonoCombAS}           & 18.44             & 22.93             & 6.31           & 28.14           \\
\cite{Hur2020SelfSupervisedMS}-ft       & 22.16             & 25.24             & 15.91          & 33.88           \\
\cite{Hur2021SelfSupervisedMM}-ft       & 22.71             & 26.51             & 13.37          & 33.09           \\
\textbf{RAFT-MSF-ft (ours)}              & \bfseries 15.96   & \bfseries 18.77   & 8.80           & \bfseries 22.50 \\

\midrule
\cite{Hur2020SelfSupervisedMS}          & 34.02             & 36.34             & 23.54           & 49.54           \\
\cite{Hur2021SelfSupervisedMM}          & 30.78             & 34.41             & 19.54           & 44.04           \\
\textbf{RAFT-MSF (ours)}                  & \bfseries 21.21   & \bfseries 27.51   & \bfseries 18.37 & \bfseries 34.98 \\
	\bottomrule
\end{tabularx}
\caption{\textbf{KITTI 2015 Scene Flow Test}: We compare our RAFT-MSF with fine-tuned or semi-supervised monocular methods (top-rows) and self-supervised method (bottom rows).}
\label{tab:testBenc}
\end{table}

Our method also demonstrates state-of-the-art monocular depth accuracy.
We evaluate our RAFT-MSF on KITTI and Eigen splits and compare with the state-of-the-art monocular depth estimation approaches in \cref{tab:depthTab}. During training and evaluating, we set the depth estimation to a fixed depth range of minimum 0m and maximum 80m. We compare the accuracy of monocular depth estimation of RAFT-MSF with recent state-of-the-art methods by using five broadly used evaluation metrics introduced in \cite{Eigen2014DepthMP}: \textit{Abs Rel}, \textit{Sq Rel}, \textit{RMSE}, \textit{RMSE log}, and threshold \textit{accuracy}.

Our method surpasses previously published methods on KITTI split for monocular depth estimation. On Eigen split, our RAFT-MSF shows better accuracy compared to \cite{Zhou_2021_ICCV} a recent monocular depth estimation method that also uses an iterative module to update the inverse depth via embedding a multi-scale feature modulation. Moreover, despite having simpler and trick-free network, we can obtain competitive accuracy with \cite{Watson2019SelfSupervisedMD} that leverages generated depth maps from the classical semi-global-matching (SGM) as proxy labels for supervision, and with \cite{Peng2021ExcavatingTP} which employs self-distillation and extensive data augmentation.

% We evaluate our RAFT-SF on KITTI and Eigen splits against multifarious monocular depth estimation approaches. Table \ref{tab:depthTab} shows that  our monocular scene flow method surpasses previously published methods on KITTI and Eigen splits for monocular depth estimation. Here, we need to highlight that our method, RAFT-SF, is not using any additional dataset or tricks. For multi-task methods \cite{Yang2018EveryPC,Liu2019UnsupervisedLO} trained on KITTI split, we outperform them by a large margin, especially in the error metrics (Abs Rel, Sq Rel, RMSE, RMSE log). Our RAFT-SF performs better than \cite{Hur2020SelfSupervisedMS} in both KITTI and Eigen splits. Moreover, RAFT-SF shows competitive performance with \cite{Watson2019SelfSupervisedMD} that leverages generated depth maps from the classical SGM as proxy labels for supervision.

\section{Conclusion}

We proposed a new self-supervised monocular scene flow method that significantly outperforms previous methods.
Our RAFT-MSF is based on RAFT and iteratively updates scene flow and disparity map using a GRU update operator.
Our contributions on disparity initialization with convolutional layers and gradient detaching strategy for upsampling layers further improve the accuracy.
Our fine-tuned version outperforms the best semi-supervised method while demonstrating 228 times faster runtime.
We hope our promising result, which demonstrates the significant accuracy boost, encourages active follow-up research on monocular scene flow estimation.

% \section{Reference}
\bibliographystyle{named}
\bibliography{ijcai22}
\end{document}